\documentclass{article}
\usepackage[T1]{fontenc}
\usepackage[utf8]{inputenc}
\usepackage{arxiv}
\usepackage{float}
\usepackage{enumitem}
\usepackage[english]{babel}
\usepackage{amssymb}
\usepackage{amsmath,mathtools}
\usepackage{stmaryrd}
\usepackage[table]{xcolor}
\usepackage{graphicx}
\usepackage{multirow}
\usepackage{cprotect}
\usepackage{bbold}
\usepackage{overpic}
\usepackage{psfrag}
\usepackage{bm}
\usepackage{lineno}
\usepackage{url}
\usepackage{physics}
\usepackage{epsfig}
\usepackage{marvosym}
\usepackage{verbatim}
\usepackage{subcaption}
\usepackage{caption}
\usepackage{hyperref}
\usepackage{ulem}
\usepackage{lmodern}
\usepackage{booktabs}
\usepackage{moresize}
\usepackage{siunitx}
\usepackage{mathtools}
\usepackage{booktabs}
\usepackage{makecell}
\usepackage{rotating,tabularx}
\usepackage{csvsimple}
\usepackage{tikz}
\usepackage{pgfplots}
\pgfplotsset{compat=1.14}
\usepgfplotslibrary{fillbetween, colormaps, colorbrewer}
\usetikzlibrary{shapes, arrows, arrows.meta, calc, positioning, decorations,
decorations.pathmorphing, decorations.text, spy}

\usepackage{etex}
\usepgfplotslibrary{external}
\tikzsetexternalprefix{tikz/}

\definecolor{blue1}		{RGB}{0,177,234}				% deep cyan
\definecolor{blue2}		{RGB}{76,200,239}				% lighter cyan
\definecolor{blue3}		{RGB}{127,215,244}				% lighter cyan
\definecolor{blue4}		{RGB}{178,231,248}				% lighter cyan
\definecolor{bluegray1}	{RGB}{0,127,167}				% deep cyan/gray mix
\definecolor{bluegray2}	{RGB}{76,165,193}				% lighter cyan/gray mix
\definecolor{bluegray3}	{RGB}{127,191,211}				% lighter cyan/gray mix
\definecolor{bluegray4}	{RGB}{178,216,228}				% lighter cyan/gray mix
\definecolor{gray1}		{RGB}{76,84,93}				% deep gray
\definecolor{gray2}		{RGB}{129,135,141}				% lighter gray
\definecolor{gray3}		{RGB}{165,169,174}				% lighter gray
\definecolor{gray4}		{RGB}{201,203,206}				% lighter gray
\definecolor{gray5}		{RGB}{230,230,230}				% lighter gray
\definecolor{gray6}		{RGB}{245,245,245}				% lighter gray
\definecolor{orange1}	{RGB}{255,126,46}				% deep orange
\definecolor{orange2}	{RGB}{255,164,108}				% lighter orange
\definecolor{orange3}	{RGB}{255,190,150}				% lighter orange
\definecolor{orange4}	{RGB}{255,216,192}				% lighter orange
\definecolor{brown1}		{RGB}{205,133,63}				% brown
\definecolor{green1}		{RGB}{0,168,107}				% green
\definecolor{green2}		{RGB}{30,198,137}				% green
\definecolor{green3}		{RGB}{60,228,167}				% green
\definecolor{green4}		{RGB}{90,255,197}				% green
\definecolor{purple1}		{RGB}{89,89,171}				% deep purple
\definecolor{purple2}		{RGB}{159,159,195}				% lighter purple
\definecolor{purple3}		{RGB}{189,189,225}				% lighter purple
\definecolor{purple4}		{RGB}{209,209,255}				% lighter purple
\definecolor{teal1}		{RGB}{0,128,128}				% deep teal
\definecolor{teal2}		{RGB}{100,168,168}				% deep teal
\definecolor{teal3}		{RGB}{130,198,198}				% deep teal
\definecolor{teal4}		{RGB}{160,228,228}				% deep teal
\definecolor{red1}		{RGB}{255,0,0}					% deep red
\definecolor{red2}		{RGB}{255,30,30}				% deep red
\definecolor{red3}		{RGB}{255,60,60}				% deep red
\definecolor{red4}		{RGB}{255,90,90}				% deep red
\definecolor{royal1}		{RGB}{2,119,189}				% royal blue
\definecolor{royal2}		{RGB}{32,149,219}				% royal blue
\definecolor{royal3}		{RGB}{62,179,249}				% royal blue
\definecolor{royal4}		{RGB}{92,209,255}				% royal blue

\usepackage[frozencache,cachedir=.]{minted}
\usepackage{tcolorbox}								% for automatic boxing and framing
\usepackage{etoolbox}								% for automatic boxing and framing

\setminted{	fontsize=\scriptsize,
      fontfamily=courier,
      fontseries=b,
      obeytabs=true,
      tabsize=2}
\setmintedinline{fontsize=\footnotesize}
\definecolor{mintedbg}{RGB}{200,200,200}
\newtcolorbox{mintedbox}{	colframe=mintedbg,
            colback=mintedbg,
            coltitle=mintedbg}
\BeforeBeginEnvironment{minted}{\begin{mintedbox}}
\AfterEndEnvironment{minted}{\end{mintedbox}}
\newcommand{\codeinline}[1]{\mintinline{cpp}|#1|}

\tikzset{arrow/.style={thick,gray2,-stealth}}
\tikzset{darrow/.style={thick,gray2,stealth-stealth}}

\renewcommand{\emph}[1]{\textit{#1}}
\setenumerate{label=\textcolor{bluegray1}{$\bm{\diamond}$},itemsep=0.0pt,topsep=5pt}

\newcommand{\ppo}{\textsc{ppo}}
\newcommand{\dqn}{\textsc{dqn}}
\newcommand{\tdt}{\textsc{td3}}
\newcommand{\sac}{\textsc{sac}}

\title{Dragonfly: a modular deep reinforcement learning library}

\def\size{7.3cm}
\author{
	\parbox{\size}{\centering J. Viquerat\thanks{Corresponding author}}\\
	MINES Paristech, CEMEF\\
	PSL - Research University\\
  \texttt{jonathan.viquerat@mines-paristech.fr}\\
\And
  \parbox{\size}{\centering P. Garnier}\\
  MINES Paristech, CEMEF\\
  PSL - Research University
\And
  \parbox{\size}{\centering A. Bateni}\\
  MINES Paristech, CEMEF\\
  PSL - Research University
\And
	\parbox{\size}{\centering E. Hachem}\\
	MINES Paristech, CEMEF\\
	PSL - Research University
}

\begin{document}
\newgeometry{left=3cm,right=3cm,top=3cm,bottom=2.5cm}
\maketitle

%%%%%%%%%%%%%%%%%%%%%%%%%%%
\begin{abstract}
  \textsc{dragonfly} is an open-source deep reinforcement learning library focused on modularity, in order to ease experimentation and developments. It relies on a \codeinline{json} serialization that allows to swap building blocks and perform parameter sweep, while minimizing code maintenance. Some of its features are specifically designed for CPU-intensive environments, such as numerical simulations. Its performance on standard agents using common benchmarks compares favorably with the literature. The library is available open-source at \url{https://github.com/jviquerat/dragonfly}.
\end{abstract}

%%%%%%%%%%%%%%%%%%%%%%%%%%%

%%%%%%%%%%%%%%%%%%%%%%%%%%%
%%%%%%%%%%%%%%%%%%%%%%%%%%%
%%%%%%%%%%%%%%%%%%%%%%%%%%%
\section{Introduction}

Deep Reinforcement Learning (DRL) has emerged as a transformative field in artificial intelligence, blending deep learning and reinforcement learning to create powerful decision-making systems \cite{sutton2018}. Recent advances have seen DRL algorithms achieving superhuman performance in complex games \cite{mnih2013, silver2017}, robotic control \cite{pinto2017}, and strategic planning \cite{kendall2018}. Still, challenges persist, and every research endeavour may require significant development and testing efforts, as well as extensive logging and results comparison.

In this contribution, we introduce \textsc{dragonfly}, a deep reinforcement learning library aimed at making prototyping and implementation of new features fast and easy. Contrarily to implementations such as \textsc{cleanrl} \cite{cleanrl} or \textsc{stable-baselines3} \cite{sb3}, which are focused on scalable, single-file implementations of agents, \textsc{dragonfly} relies on a highly-modular pattern. Its performance level is ensured by constant performance evaluation of the agents against standard benchmarks, making it a reliable production tool. The library relies on \textsc{tensorflow} 2 as a backend for the design of neural network architectures.

%%%%%%%%%%%%%%%%%%%%%%%%%%%
%%%%%%%%%%%%%%%%%%%%%%%%%%%
%%%%%%%%%%%%%%%%%%%%%%%%%%%
\section{Library architecture}

%%%%%%%%%%%%%%%%%%%%%%%%%%%
%%%%%%%%%%%%%%%%%%%%%%%%%%%
\subsection{General architecture}

The global architecture of the library is summed up in figure \ref{fig:architecture}. The training is driven by a \codeinline{.json} file that provides all the informations required to set the different classes and train the agent. The training procedure is handled by a \codeinline{trainer} object, which in turn initializes the \codeinline{environment} and the \codeinline{agent}, as well as several other objects helpful for the logging, rendering, etc. These elements rely on lower level objects such as networks, optimizers and losses.

\input{fig/architecture}

%%%%%%%%%%%%%%%%%%%%%%%%%%%
%%%%%%%%%%%%%%%%%%%%%%%%%%%
\subsection{Implementation details}

The backbone of the present library is composed of three components: (i) a \codeinline{.json} serialization of the different components, (ii) the use of \codeinline{SimpleNamespace} instances to store the parameters of each object, and (iii) a factory pattern.

The \codeinline{.json} parsing relies on the standard \codeinline{json} module, and exploits the \codeinline{SimpleNamespace} iterable-based constructor to provide a lightweight and easy-to-use class-like representation of the serialized inputs. Finally, the \codeinline{factory} class allows to generate objects of a desired type using a string representation:

\begin{minted}{python}
class factory:
   def __init__(self):
      self.keys = {}

   def register(self, key, creator):
      self.keys[key] = creator

   def create(self, key, **kwargs):
      creator = self.keys.get(key)
      if not creator:
         try:
            raise ValueError(key)
         except ValueError:
            error("factory", "create", "Unknown key provided: "+key)
            raise

      return creator(**kwargs)
\end{minted}

This pattern adds a lot of flexibility to the library, as swapping building blocks within a given algorithm only requires to modify the input \codeinline{.json} configuration file. For each object type, (agents, losses, etc), a factory is instantiated, and the different corresponding classes are registered using the \codeinline{register} member with a key/value pair:

\begin{minted}{python}
agent_factory = factory()

agent_factory.register("a2c",  a2c)
agent_factory.register("ppo",  ppo)
agent_factory.register("dqn",  dqn)
agent_factory.register("ddpg", ddpg)
agent_factory.register("td3",  td3)
agent_factory.register("sac",  sac)
\end{minted}

Then, generating an object of a given type (here an agent) from a string representation is performed using the \codeinline{create} member:

\begin{minted}{python}
  agent = agent_factory.create(agent_parameters.type,
                               spaces         = environment.spaces,
                               n_environments = mpi.size,
                               memory_size    = memory_size,
                               parameters     = agent_parameters)
\end{minted}

%%%%%%%%%%%%%%%%%%%%%%%%%%%
%%%%%%%%%%%%%%%%%%%%%%%%%%%
%%%%%%%%%%%%%%%%%%%%%%%%%%%
\section{Features}

%%%%%%%%%%%%%%%%%%%%%%%%%%%
%%%%%%%%%%%%%%%%%%%%%%%%%%%
\subsection{Agents}

Standard agents, such as \textsc{ppo}, \textsc{dqn}, \textsc{ddpg}, \textsc{td3} or \textsc{sac}, are implemented in the library. The agent training is based on different \codeinline{trainer} types, based on the on-policy or off-policy nature of the agent. The \codeinline{trainer} is a cornerstone structure of the library, as it is used to instantiate the agent and the environment, as well as several other objects used to train the agent and log its performances. As for some other structures of the library, this wrapper provides another layer of flexibility in the exploration of DRL algorithms.

Each agent generates some \codeinline{policy} and \codeinline{value} instances, the latter spawning adequate \codeinline{network}, \codeinline{loss} and \codeinline{optimizer} instances based on the description of the \codeinline{json} serialization. Additional types may be required, such as \codeinline{return} and \codeinline{termination} types, depending on the agent type.

%%%%%%%%%%%%%%%%%%%%%%%%%%%
%%%%%%%%%%%%%%%%%%%%%%%%%%%
\subsection{Environment}

The \codeinline{environment} class consists in a wrapper around (i) a \codeinline{spaces} class, and (ii) a set of \codeinline{worker} instances. The \codeinline{spaces} instance is devoted to handling the relation with the actual environment, determining the size and shape of actions and observations, and applying transformations to the latter if needed. Each \codeinline{worker} instance wraps an actual environment instance, and interacts with the upper layer. Doing so, considering a single or multiple environments is made transparent to the \codeinline{agent} instance. The transformations applied to the actions and observations can be directly parameterized through the \codeinline{.json} file, and optional arguments can also be passed transparently through the different layers to the actual environment.

%%%%%%%%%%%%%%%%%%%%%%%%%%%
%%%%%%%%%%%%%%%%%%%%%%%%%%%
\subsection{Buffer}

Buffering is made flexible using multiple layers of buffers to store data on-the-fly from the parallel environments. Right before training, parallel informations are collected from the parallel buffer and reshaped to be stored in a larger memory buffer, making them avaiable for sampling during the training phase. To limit the memory footprint, a ring buffer structure is used. The buffer structures are initialized using dict structures, so the same buffer class can be initialized using different lists of key/size pairs, and therefore be used with different agents.

%%%%%%%%%%%%%%%%%%%%%%%%%%%
%%%%%%%%%%%%%%%%%%%%%%%%%%%
\subsection{Bootstrapping for parallel environments}

In the context of CPU-intensive environments (for example, when the environment consists in a finite element resolution of a physical problem), the use of parallel environments can rapidly become a necessity. Yet, for methods such as \textsc{ppo}, the on-policiness assumption may be broken when using too many simultaneous environments (for example, if the agent update requires four full trajectories, the on-policiness assumption is broken if more than four environments are unrolled simultaneously). In this context, a bootstrapping termination technique can be used to mimic the on-policiness assumption beyond the theoretical limit. As presented in \cite{viquerat2023}, and reproduced here in figure \ref{fig:parallel}, this simple method allows to use $4$ to $8$  times more parallel environments than the vanilla approach. For additional details, the reader is referred to \cite{viquerat2023}.

\input{fig/parallel}

%%%%%%%%%%%%%%%%%%%%%%%%%%%
%%%%%%%%%%%%%%%%%%%%%%%%%%%
\subsection{State representation learning}

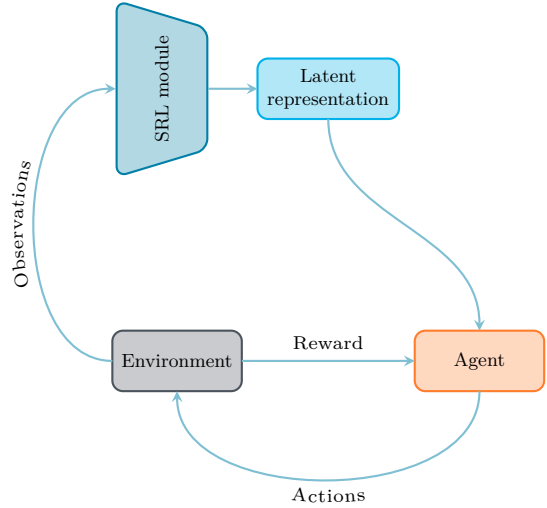
\begin{figure}
  \centering
  \def\myscale{0.8}
  \begin{tikzpicture}[	scale=\myscale,
      arrow/.style={thick, color=bluegray3, rounded corners}]

    %%%%% autoencoder
    \node[	scale=\myscale,
      trapezium, trapezium angle=45, rotate=90, trapezium stretches body, shape border rotate=180,
      minimum height=1.5cm, minimum width=2mm, rounded corners,
      draw=bluegray1, fill=bluegray4, thick]
    (enc) at (-0.25,4.5) {\footnotesize SRL module};
    \node[	scale=\myscale,
      rectangle, rounded corners, draw=blue1, fill=blue4, thick,
      text width=2.2cm, minimum height=1cm, text centered,
      inner sep=2pt, outer sep=0pt]
    (feat) at (2.5,4.5) {\footnotesize Latent\\ representation};

    %%%%% nodes
    \node[	scale=\myscale,
      rectangle, rounded corners, draw=gray1, fill=gray4, thick,
      text width=2cm, minimum height=1cm, text centered,
      inner sep=2pt, outer sep=0pt]
    (env) at (0,0) {\footnotesize Environment};
    \node[	scale=\myscale,
      rectangle, rounded corners, draw=orange1, fill=orange4, thick,
      text width=2cm, minimum height=1cm, text centered,
      inner sep=2pt, outer sep=0pt]
    (agent) at (5,0) {\footnotesize Agent};

    %%%%% arrows
    \draw[-stealth,arrow] (env.east) -- (agent.west);
    \draw[-stealth,arrow] (agent.south) to [out=-90,in=-90] (env.south);

    \draw[-stealth,arrow] (env.west) to [out=180,in=180] (enc.north);
    \draw[-stealth,arrow] (enc.south) -- (feat.west);
    \draw[-stealth,arrow] (feat.south) to [out=-90,in=90] (agent.north);

    %%%%% arrow writings
    \draw[draw=none,postaction={decorate,decoration={raise=1ex,text along path,text align=center,text={|\scriptsize|Reward}}}] (env.east) -- (agent.west);
    \draw[draw=none,postaction={decorate,decoration={raise=-2ex,text along path,text align=center,text={|\scriptsize|Actions}}}] (env.south) to [out=-90,in=-90] (agent.south);
    \draw[draw=none,postaction={decorate,decoration={raise=1ex,text along path,text align=center,text={|\scriptsize|Observations}}}] (env.west) to [out=180,in=180] (enc.north);

  \end{tikzpicture}
  \caption{\textbf{State representation learning setup in the DRL learning loop.} The traditional observation feedback is replaced by a modified feedback loop, in which observations are transformed in a latent representation before being fed to the agent.}
  \label{fig:srl}
\end{figure}

Representation learning or feature learning refers to methods providing significative representations of raw data fulfilling specified goals. This field encompasses methods such as principal component analysis (\textsc{pca}), clustering techniques, auto-encoders (\textsc{ae}), etc. In most DRL-based control cases, the observations provided by the environment may be noisy, incomplete, high-dimensional and/or highly correlated, which may lead to subpar control performance from the agent. In this context, state representation learning aims to process these observations upstream of the DRL agent, by learning a modified or a low-dimensional representation of the original data, called latent representation (see figure \ref{fig:srl}).

In its current state, the library integrates a \textsc{pca} and an auto-encoder modules, that can be seamlessly integrated to existing agents. The coupling of an agent with state representation learning can be split in two steps. First, a warmup phase is defined during which random actions are taken by the agent, while the observations are collected for the training of the SRL module. No update of the agent is performed during this phase. Once a pre-defined amount of samples have been collected, the SRL module is updated, and the training phase of the agent starts. During this second phase, the agent is fed with SRL-processed observations as input. This on-the-fly process is easy to use, as it does not require any pre-processing of the environment, nor any third-party module to train the SRL module. Of course, the trained parameters of the SRL module can be saved and re-used for future trainings if needed.

\begin{figure}
\centering
%%%%%%%%%%%%
\begin{subfigure}[t]{.35\textwidth}
\centering
\begin{tikzpicture}[	trim axis left, trim axis right, font=\scriptsize,
        upper/.style={	name path=upper, smooth, draw=none},
        lower/.style={	name path=lower, smooth, draw=none},]
  \begin{axis}[	xmin=0, xmax=1000000, scale=0.7,
        ymin=-5, ymax=0,
        scaled x ticks=false, scaled y ticks=false,
        xtick={0,200000,400000,600000,800000,1000000},
        xticklabels={$0$,$200k$,$400k$,$600k$,$800k$,$1000k$},
        legend cell align=left, legend pos=north west,
        legend style={nodes={scale=0.8, transform shape}},
        every tick label/.append style={font=\scriptsize},
        grid=major, xlabel=transitions, ylabel=score]

    \legend{\textsc{ppo}(1000), \textsc{ppo-pca}(300), \textsc{ppo-pca}(50), \textsc{ppo-pca}(10)}

    \addplot[upper, forget plot]        table[x index=0,y index=7] {fig/srl/ppo_20_jets.dat};
    \addplot[lower, forget plot]        table[x index=0,y index=6] {fig/srl/ppo_20_jets.dat};
    \addplot[fill=gray3, opacity=0.5, forget plot]  fill between[of=upper and lower];
    \addplot[draw=gray1, thick, smooth]       table[x index=0,y index=5] {fig/srl/ppo_20_jets.dat};

    \addplot [upper, forget plot]         table[x index=0,y index=7] {fig/srl/ppo_pca_dim_300.dat};
    \addplot [lower, forget plot]         table[x index=0,y index=6] {fig/srl/ppo_pca_dim_300.dat};
    \addplot [fill=red3, opacity=0.5, forget plot]  fill between[of=upper and lower];
    \addplot[draw=red1, thick, smooth]      table[x index=0,y index=5] {fig/srl/ppo_pca_dim_300.dat};

    \addplot [upper, forget plot]         table[x index=0,y index=7] {fig/srl/ppo_pca_dim_50.dat};
    \addplot [lower, forget plot]         table[x index=0,y index=6] {fig/srl/ppo_pca_dim_50.dat};
    \addplot [fill=teal3, opacity=0.5, forget plot]   fill between[of=upper and lower];
    \addplot[draw=teal1, thick, smooth]       table[x index=0,y index=5] {fig/srl/ppo_pca_dim_50.dat};

    \addplot [upper, forget plot]         table[x index=0,y index=7] {fig/srl/ppo_pca_dim_10.dat};
    \addplot [lower, forget plot]         table[x index=0,y index=6] {fig/srl/ppo_pca_dim_10.dat};
    \addplot [fill=green3, opacity=0.5, forget plot]  fill between[of=upper and lower];
    \addplot[draw=green1, thick, smooth]    table[x index=0,y index=5] {fig/srl/ppo_pca_dim_10.dat};

  \end{axis}
\end{tikzpicture}
\caption{Score curves}
\label{fig:shkadov_pca_score}
\end{subfigure} \qquad \qquad
%%%%%%%%%%%%
%%%%%%%%%%%%
\begin{subfigure}[t]{.35\textwidth}
  \centering
  \begin{tikzpicture}[	trim axis left, trim axis right, font=\scriptsize]
	  \begin{semilogxaxis}[	xmin=0, xmax=1000, scale=0.7,
				ymin=0, ymax=1,
				xtick={10,100,1000},
				xticklabels={10,100,1000},
				scaled x ticks=false, scaled y ticks=false,
				every tick label/.append style={font=\scriptsize},
				grid=major, xlabel=latent space dimension, ylabel=explained variance]
		  \addplot[mark=*,mark options={fill=gray2}, thick, gray1] coordinates {(10,0.35) (50,0.82) (100,0.97) (300,0.999) (1000,1)};
	  \end{semilogxaxis}
  \end{tikzpicture}
  \caption{Explained variance}
  \label{fig:shkadov_pca_explained_variance}
\end{subfigure}
%%%%%%%%%%%%

\medskip
\medskip

\begin{subfigure}[t]{.8\textwidth}
	\includegraphics[height=.5cm,width=\textwidth]{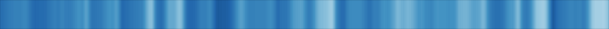}
  \includegraphics[height=.5cm,width=.3\textwidth]{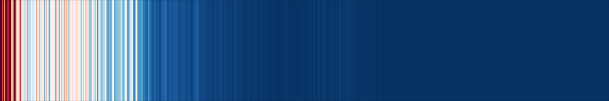}
  \caption{Hotmap of the observations standard deviation averaged over one episode without (top) and with (bottom) the \textsc{pca} module for a latent space dimension of $300$.}
  \label{fig:shkadov_pca_std}
\end{subfigure}

%%%%%%%%%%%%
\caption{\textbf{Comparing} \textsc{ppo} \textbf{with} \textsc{ppo-pca} \textbf{on the} \codeinline{shkadov-v0} \textbf{ environment}. (Top left) The \textsc{ppo-pca} overperforms the standard \textsc{ppo} algorithm for an observation vector of size \num{1000} corresponding to the entire information of the considered environment. Performance varies based on the chose latent space dimension. (Top right) Adequate latent space dimension can be found by evaluating the explained variance of the \textsc{pca} representation chosen, the optimal corresponding to the lowest dimension for which the explained variance remains close to $1$. (Bottom) The hotmap of the observations standard deviation averaged over one episode without and with the \textsc{pca} module shows that the \textsc{pca} module generates observations with high variance.}
\label{fig:shkadov_ppo_pca}
\end{figure}

In figure \ref{fig:shkadov_pca_score}, a performance comparison with and without the \textsc{srl} module is shown on the \codeinline{shkadov-v0} environment from the \textsc{beacon} benchmark \cite{beacon}. The entire set of observations is provided to the agent as a vector of size \num{1000}. In the regular case, the \textsc{ppo} agent directly processes the input vector, while in the \textsc{srl} case, a \textsc{pca} representation of these observations is built on-the-fly, and then fed to the agent. One sees that the \textsc{ppo} $+$ \textsc{srl} agent with a latent space of size $300$ overperforms the standard \textsc{ppo} agent. It is also observed that too low latent space dimensions lead to sub-par performance, or even no learning at all. The optimal latent space dimension is obtained by looking for the lowest dimension for which the explained variance remains close to $1$, as is shown in figure \ref{fig:shkadov_pca_explained_variance}. To further illustrate the interest of this method, a hotmap of the observations standard deviation averaged over one episode is shown in figure \ref{fig:shkadov_pca_std}. As can be seen, the \textsc{pca} method naturally builds latent features with large variance, which can be advantageous in the context of DRL as it leads to a condensed representation of the important variations in the observations of the environment, while filtering out the low-variation features.

%%%%%%%%%%%%
%%%%%%%%%%%%
%%%%%%%%%%%%
%%%%%%%%%%%%

%%%%%%%%%%%%%%%%%%%%%%%%%%%
%%%%%%%%%%%%%%%%%%%%%%%%%%%
\subsection{Separable environments}

Learning a proper control strategy can become nearly impossible in the case of high-dimensional action spaces, leading to excessive sample requirements. Yet, some environments presenting adequate separability can be reformulated as problems of lower action dimensionality, leading to an efficient learning. This approach was presented in \cite{belus2019}, and is implemented in the library as a specific \codeinline{trainer} class, handling the mapping from $1$ environment with $n_\text{act}$ actions to $n_\text{act}$ environments with $1$ action. Doing so, the problem is simplified by (i) reducing the observation space dimensionality and (ii) multiplying by $n_\text{act}$ the number of available samples to train the agent. This results in a significant learning speedup, especially for large action spaces.

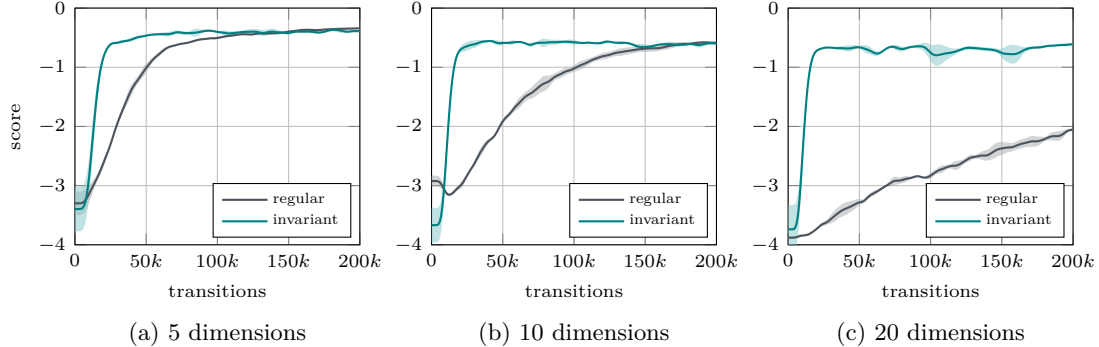
\begin{figure}
\centering
%%%%%%%%%%%%
\begin{subfigure}[t]{.25\textwidth}
  \centering
  \begin{tikzpicture}[	trim axis left, trim axis right, font=\scriptsize,
          upper/.style={	name path=upper, smooth, draw=none},
          lower/.style={	name path=lower, smooth, draw=none},]
    \begin{axis}[	xmin=0, xmax=200000, scale=0.55,
          ymin=-4, ymax=0,
          scaled x ticks=false,
          xtick={0,50000,100000,150000,200000},
          xticklabels={$0$,$50k$,$100k$,$150k$,$200k$},
          ytick={-4,-3,-2,-1,0},
          yticklabels={$-4$,$-3$,$-2$,$-1$,$0$},
          legend cell align=left, legend pos=south east,
          legend style={nodes={scale=0.8, transform shape}},
          every tick label/.append style={font=\scriptsize},
          grid=major, xlabel=transitions, ylabel=score]

      \legend{regular, invariant}

      \addplot [upper, forget plot]         table[x index=0,y index=7] {fig/invariant/5_jets.dat};
      \addplot [lower, forget plot]         table[x index=0,y index=6] {fig/invariant/5_jets.dat};
      \addplot [fill=gray3, opacity=0.5, forget plot]   fill between[of=upper and lower ];
      \addplot[draw=gray1, thick, smooth]   table[x index=0,y index=5] {fig/invariant/5_jets.dat};

      \addplot [upper, forget plot]         table[x index=0,y index=7] {fig/invariant/5_jets_invariant.dat};
      \addplot [lower, forget plot]         table[x index=0,y index=6] {fig/invariant/5_jets_invariant.dat};
      \addplot [fill=teal3, opacity=0.5, forget plot]   fill between[of=upper and lower];
      \addplot[draw=teal1, thick, smooth]   table[x index=0,y index=5] {fig/invariant/5_jets_invariant.dat};

    \end{axis}
  \end{tikzpicture}
  \caption{$5$ dimensions}
  \label{fig:shkadov_5_jets}
\end{subfigure} \qquad
%%%%%%%%%%%%
\begin{subfigure}[t]{.25\textwidth}
  \centering
  \begin{tikzpicture}[	trim axis left, trim axis right, font=\scriptsize,
          upper/.style={	name path=upper, smooth, draw=none},
          lower/.style={	name path=lower, smooth, draw=none},]
    \begin{axis}[	xmin=0, xmax=200000, scale=0.55,
        ymin=-4, ymax=0,
        scaled x ticks=false,
        xtick={0,50000,100000,150000,200000},
        xticklabels={$0$,$50k$,$100k$,$150k$,$200k$},
        ytick={-4,-3,-2,-1,0},
        yticklabels={$-4$,$-3$,$-2$,$-1$,$0$},
        legend cell align=left, legend pos=south east,
        legend style={nodes={scale=0.8, transform shape}},
        every tick label/.append style={font=\scriptsize},
        grid=major, xlabel=transitions, ylabel={}]

      \legend{regular, invariant}

      \addplot [upper, forget plot]         table[x index=0,y index=7] {fig/invariant/10_jets.dat};
      \addplot [lower, forget plot]         table[x index=0,y index=6] {fig/invariant/10_jets.dat};
      \addplot [fill=gray3, opacity=0.5, forget plot]   fill between[of=upper and lower];
      \addplot[draw=gray1, thick, smooth]   table[x index=0,y index=5] {fig/invariant/10_jets.dat};

      \addplot [upper, forget plot]         table[x index=0,y index=7] {fig/invariant/10_jets_invariant.dat};
      \addplot [lower, forget plot]         table[x index=0,y index=6] {fig/invariant/10_jets_invariant.dat};
      \addplot [fill=teal3, opacity=0.5, forget plot]   fill between[of=upper and lower];
      \addplot[draw=teal1, thick, smooth]   table[x index=0,y index=5] {fig/invariant/10_jets_invariant.dat};

    \end{axis}
  \end{tikzpicture}
  \caption{$10$ dimensions}
  \label{fig:shkadov_10_jets}
\end{subfigure} \qquad
%%%%%%%%%%%%
\begin{subfigure}[t]{.25\textwidth}
  \centering
  \begin{tikzpicture}[	trim axis left, trim axis right, font=\scriptsize,
          upper/.style={	name path=upper, smooth, draw=none},
          lower/.style={	name path=lower, smooth, draw=none},]
    \begin{axis}[	xmin=0, xmax=200000, scale=0.55,
        ymin=-4, ymax=0,
        scaled x ticks=false,
        xtick={0,50000,100000,150000,200000},
        xticklabels={$0$,$50k$,$100k$,$150k$,$200k$},
        ytick={-4,-3,-2,-1,0},
        yticklabels={$-4$,$-3$,$-2$,$-1$,$0$},
        legend cell align=left, legend pos=south east,
        legend style={nodes={scale=0.8, transform shape}},
        every tick label/.append style={font=\scriptsize},
        grid=major, xlabel=transitions, ylabel={}]

      \legend{regular, invariant}

      \addplot [upper, forget plot]         table[x index=0,y index=7] {fig/invariant/20_jets.dat};
      \addplot [lower, forget plot]         table[x index=0,y index=6] {fig/invariant/20_jets.dat};
      \addplot [fill=gray3, opacity=0.5, forget plot]   fill between[of=upper and lower];
      \addplot[draw=gray1, thick, smooth]   table[x index=0,y index=5] {fig/invariant/20_jets.dat};

      \addplot [upper, forget plot]         table[x index=0,y index=7] {fig/invariant/20_jets_invariant.dat};
      \addplot [lower, forget plot]         table[x index=0,y index=6] {fig/invariant/20_jets_invariant.dat};
      \addplot [fill=teal3, opacity=0.5, forget plot]   fill between[of=upper and lower];
      \addplot[draw=teal1, thick, smooth]   table[x index=0,y index=5] {fig/invariant/20_jets_invariant.dat};

    \end{axis}
  \end{tikzpicture}
  \caption{$20$ dimensions}
  \label{fig:shkadov_20_jets}
\end{subfigure}
%%%%%%%%%%%%
\caption{\textbf{Comparing traditional} \textsc{ppo} \textbf{with separable} \textsc{ppo} on the \codeinline{shkadov-v0} environment with $5-$, $10-$ and $20-$ dimensional action spaces. Note that the reward is not computed in the same manner as in \cite{belus2019}.}
\label{fig:shkadov_invariant}
\end{figure}

In figure \ref{fig:shkadov_invariant}, we reproduce the main results from \cite{belus2019} by comparing the score curves with and without the separability feature on the \codeinline{shkadov-v0} environment, with $5$-, $10$- and $20$-dimensional action spaces (for more details, please refer to \cite{beacon}). As can be observed, the performance of the separable formulation remains steady even for high action dimensionality.

%%%%%%%%%%%%
%%%%%%%%%%%%
%%%%%%%%%%%%
%%%%%%%%%%%%

%%%%%%%%%%%%%%%%%%%%%%%%%%%
%%%%%%%%%%%%%%%%%%%%%%%%%%%
%%%%%%%%%%%%%%%%%%%%%%%%%%%
\section{Performance and benchmarks}

The agents implementations are tested on three different litterature benchmarks: (i) the \textsc{gymnasium} environments \cite{gymnasium}, (ii) the \textsc{mujoco} benchmark \cite{mujoco}, and (iii) the \textsc{beacon} benchmark \cite{beacon}, the latter being a benchmark library dedicated to flow control problems. Results are presented respectively in figures \ref{fig:gym}, \ref{fig:mujoco} and \ref{fig:beacon}.

\input{fig/gym}

\input{fig/mujoco}

\input{fig/beacon}

%%%%%%%%%%%%%%%%%%%%%%%%%%%
\section{Conclusion}

The present contribution presented the \textsc{dragonfly} library, its modular construction pattern and several of its interesting features. The performance levels of the different agents was also assessed on well-known literature benchmarks. The library is available open-source at \url{https://github.com/jviquerat/dragonfly}.

%%%%%%%%%%%%%%%%%%%%%%%%%%%
\appendix
\section{Acknowledgements}

Funded/Co-funded by the European Union (ERC, CURE, 101045042). Views and opinions expressed are however those of the author(s) only and do not necessarily reflect those of the European Union or the European Research Council. Neither the European Union nor the granting authority can be held responsible for them.

\bibliographystyle{unsrt}
\bibliography{refs}

\begin{thebibliography}{10}

\bibitem{sutton2018}
R.~S. Sutton and A.~G. Barto.
\newblock {\em Reinforcement {L}earning: {A}n {I}ntroduction}.
\newblock MIT Press, Cambridge, MA, 2018.

\bibitem{mnih2013}
V.~Mnih, K.~Kavukcuoglu, D.~Silver, A.~Graves, I.~Antonoglou, D.~Wierstra, and
  M.~Riedmiller.
\newblock Playing {Atari} with deep reinforcement learning.
\newblock {\em arXiv preprint arXiv:1312.5602}, 2013.

\bibitem{silver2017}
D.~Silver, J.~Schrittwieser, K.~Simonyan, I.~Antonoglou, A.~Huang, A.~Guez,
  T.~Hubert, L.~Baker, M.~Lai, A.~Bolton, Y.~Chen, T.~Lillicrap, F.~Hui,
  L.~Sifre, G.~van~den Driessche, T.~Graepel, and D.~Hassabis.
\newblock Mastering the game of {G}o without human knowledge.
\newblock {\em Nature}, 550, 2017.

\bibitem{pinto2017}
L.~Pinto, M.~Andrychowicz, P.~Welinder, W.~Zaremba, and P.~Abbeel.
\newblock Asymmetric actor critic for image-based robot learning.
\newblock {\em arXiv preprint arXiv:1710.06542}, 2017.

\bibitem{kendall2018}
A.~Kendall, J.~Hawke, D.~Janz, P.~Mazur, D.~Reda, J.-M. Allen, V.-D. Lam,
  A.~Bewley, and A.~Shah.
\newblock Learning to drive in a day.
\newblock {\em arXiv preprint arXiv:1807.00412}, 2018.

\bibitem{cleanrl}
S.~Huang, R.~F.~J. Dossa, C.~Ye, and J.~Braga.
\newblock Cleanrl: High-quality single-file implementations of deep
  reinforcement learning algorithms, 2021.

\bibitem{sb3}
A.~Raffin, A.~Hill, A.~Gleave, A.~Kanervisto, M.~Ernestus, and N.~Dormann.
\newblock Stable-baselines3: Reliable reinforcement learning implementations.
\newblock {\em Journal of Machine Learning Research}, 22(268):1--8, 2021.

\bibitem{viquerat2023}
J.~Viquerat and E.~Hachem.
\newblock Parallel bootstrap-based on-policy deep reinforcement learning for
  continuous flow control applications, 2023.

\bibitem{beacon}
J.~Viquerat, P.~Meliga, P.~Jeken-Rico, and E.~Hachem.
\newblock Beacon, a lightweight deep reinforcement learning benchmark library
  for flow control.
\newblock {\em Applied Sciences}, 14(9), 2024.

\bibitem{belus2019}
V.~Belus, J.~Rabault, J.~Viquerat, Z.~Che, E.~Hachem, and U.~Reglade.
\newblock Exploiting locality and translational invariance to design effective
  deep reinforcement learning control of the 1-dimensional unstable falling
  liquid film.
\newblock {\em AIP Advances}, 9:125014, 2019.

\bibitem{gymnasium}
M.~Towers, A.~Kwiatkowski, J.~Terry, J.~U. Balis, G.~De Cola, T.~Deleu,
  M.~Goulão, A.~Kallinteris, M.~Krimmel, A.~KG, R.~Perez-Vicente, A.~Pierré,
  S.~Schulhoff, J.~J. Tai, H.~Tan, and O.~G. Younis.
\newblock Gymnasium: A standard interface for reinforcement learning
  environments, 2024.

\bibitem{mujoco}
E.~Todorov, T.~Erez, and Y.~Tassa.
\newblock Mujoco: A physics engine for model-based control.
\newblock In {\em 2012 IEEE/RSJ International Conference on Intelligent Robots
  and Systems}, pages 5026--5033. IEEE, 2012.

\end{thebibliography}

\end{document}